\PassOptionsToPackage{table}{xcolor}
\documentclass[10pt,twocolumn,letterpaper]{article}
\usepackage[pagenumbers]{cvpr}
\usepackage[OT1]{fontenc}
\usepackage{cuted}
\usepackage{caption}
\usepackage{booktabs}
\usepackage{multirow}
\usepackage{graphicx}
\usepackage{amsmath}
\usepackage{amssymb}
\usepackage{bm}
\usepackage{xspace}
\usepackage{makecell}

\definecolor{cvprblue}{rgb}{0.21,0.49,0.74}
\usepackage[pagebackref,breaklinks,colorlinks,allcolors=cvprblue]{hyperref}
\def\confName{CVPR}
\def\confYear{2027}

\title{FireDesmoke: Fire Smoke Removal via Real Smoke Transfer and Semantic Rectification}
\author{Chih-Chung Hsu \qquad Rong-Lin Jian\\
National Yang Ming Chiao Tung University, Taiwan\\
{\tt\small chihchung@nycu.edu.tw}}

\newcommand{\FDMcolor}{\ifdefined\FDMshowrev\color{red}\else\color{black}\fi}

\definecolor{rankred}{HTML}{B2182B}
\definecolor{rankblue}{HTML}{2166AC}
\newcommand{\rankfirst}[1]{{\color{rankred}\bfseries #1}}
\newcommand{\ranksecond}[1]{{\color{rankblue}\underline{#1}}}

\newcommand{\SRCTAB}{S11}\newcommand{\SMOKETAB}{S7}\newcommand{\HAZETAB}{S9}\newcommand{\NEXT}{116}\newcommand{\HALLFIG}{S4}\newcommand{\PERSONTAB}{S16}\newcommand{\LOCFIG}{S5}
\begin{document}
\maketitle

\begin{abstract}
Fire smoke removal must restore dense and medium smoke regions while preserving the clear (smoke-free) content. Existing methods either leave the smoke or alter the visible content, and generative models may hallucinate people or remove flames. Real fire smoke is spatially local: the degradation and its corruption of semantic features are both concentrated in the smoke regions. We propose FireDesmoke, built on this locality: Real Smoke Transfer calibrates smoke fields on real-fire recordings and transfers their spatial layout onto clean images to synthesize training pairs, and Semantic Rectification uses pixel features to rectify the tokens of a frozen foundation encoder where smoke corrupts them. With matched training budgets, replacing the transferred pairs with SmokeBench pairs lowers dense-smoke PSNR by 0.74 dB, and removing Semantic Rectification leaves clear regions 1.31 dB below the input. On what is, to the best of our knowledge, the first paired dataset of real fire smoke for image restoration, with 1,500 test frames from a held-out viewpoint, FireDesmoke improves dense- and medium-smoke PSNR by 3.73 and 2.63 dB while keeping clear regions at the input level ($+$0.41 dB); no other compared method gains at least 1 dB in dense smoke without lowering clear-region PSNR, and no hallucinated person is found in its outputs. It achieves state-of-the-art overall results on this benchmark and, trained on SmokeBench, the highest PSNR with the second-best SSIM and LPIPS.
\end{abstract}

\begin{figure}[t]\FDMcolor 
\centering
\includegraphics[width=\columnwidth]{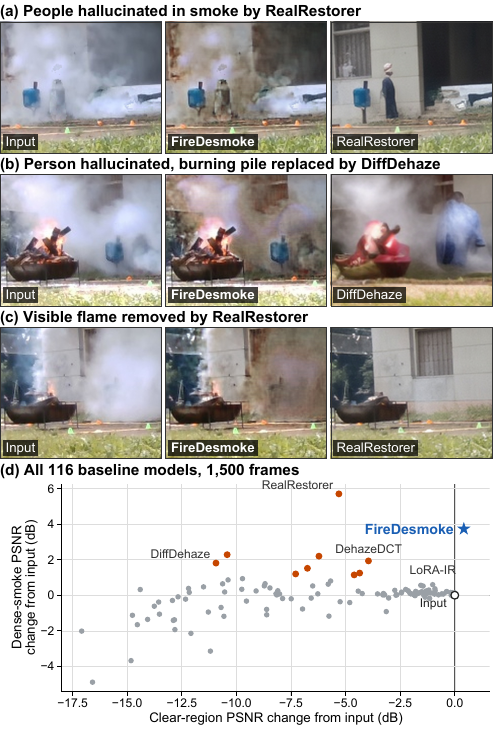}
\caption{\FDMcolor \textbf{Reliability of fire smoke removal on our test set (held-out viewpoint, zero-shot for all methods).} (a, b) In selected default-setting outputs, RealRestorer~\cite{rw_l202} and DiffDehaze~\cite{hazegen} hallucinate people in smoke where the input shows none, and DiffDehaze replaces the burning pile. (c) RealRestorer removes a visible flame. FireDesmoke adds no person to these frames and keeps the flame. (d) Trade-off over all 1,500 frames: change from the input in dense-smoke and clear-region RGB PSNR for all \NEXT{} baseline models and FireDesmoke; higher is better on both axes. Baselines that gain at least 1 dB in dense smoke (orange) lose at least 3.9 dB in clear regions, while FireDesmoke gains 3.73 dB with clear regions at +0.41 dB. Labeled models: Table S2; larger crops and settings: Fig. \HALLFIG{}, Sec. S9.}
\label{fig:teaser}
\end{figure}

\section{Introduction}
\label{sec:intro}
In fire scenes, firefighters and perception models rely on images that smoke and flames already degrade. Smoke removal helps only if its output is reliable: it should restore the regions covered by smoke and preserve the content that is already visible. A method that violates the second requirement replaces information in the input with its own, from shifted colors, surfaces, and boundaries to replaced objects~\cite{rw_x_llbench} and, in the worst case, a hallucinated person or a removed flame. Existing methods do not satisfy both: among \NEXT{} baseline models, those that remove dense smoke alter the clear regions, and those that preserve the clear regions leave the smoke (\cref{fig:teaser}d).

Single-image desmoking~\cite{smokebench} restores scenes degraded by smoke~\cite{figlib,firestereo,subtmrs}. Unlike the homogeneous haze of dehazing benchmarks~\cite{reside,ohaze,ihaze,densehaze}, fire smoke is spatially non-homogeneous and semi-transparent: dense smoke covers part of the frame, its density decreases continuously toward clear surroundings (\cref{fig:qual}), and its distribution changes with the wind. Restoration must therefore be spatially adaptive, removing smoke where it is present and preserving clear content elsewhere. A binary smoke mask is insufficient: semi-transparent smoke has no sharp boundary. We instead learn this behavior in a single mask-free network, which requires training data that reflect where real smoke occurs.

\emph{Paired data of real fire smoke are very difficult to collect.} Real buildings that can be set on fire, with a fire engine and firefighters on site, are available only at a fire department's search-and-rescue training ground. The application takes about two months and, once approved, about one more to schedule; the site must be in good condition and the weather dry; and the fire department decides the fuel, the camera positions, and when the fire may grow. One authorized day provided about 40 minutes of outdoor fire. The cameras must also remain fixed so that clear frames of the same view can serve as references while firelight, water spray, and moving staff change the scene. Hence, to the best of our knowledge, no paired dataset of real fire smoke for restoration has existed, and scene diversity cannot come from the fire itself: we take from the recordings only what a real fire alone can provide, its smoke fields (effective transmission with its atmospheric light and exposure gain), and take the scene content from public clean images, so each smoke field is reused across many scenes.

\emph{Training data must reproduce where the degradation occurs.} A transmission histogram describes the density of the smoke but not its spatial layout, a coherent smoke region with clear surroundings. In our calibrated smoke fields, 68\% of the degradation (input--target difference) lies inside the smoke region (transmission $T{<}0.5$), which covers 23\% of a training window (Sec. S8); training smoke must preserve this layout together with its density.

\emph{Semantic priors are least reliable where restoration is most needed.} Features of a foundation encoder guide restoration well where the degradation leaves them intact. Under fire smoke, their reliability is strongly local: at the deepest encoder block, the tokens of the smoky input (input tokens) have a cosine similarity of only 0.54 to those of the clear reference in dense smoke, vs.\ 0.98 in clear regions. Used without correction, these tokens lead our model to alter clear content even far from the smoke (\cref{sec:mech}).

We propose FireDesmoke, a mask-free framework for single-image fire smoke removal in which both the training data and the semantic priors follow the spatial distribution of real smoke (\cref{fig:pipeline}). \emph{Real Smoke Transfer} analytically calibrates smoke fields on fixed-view real-fire recordings and transfers them, with their spatial layout, onto public clean images; the same transmission balances the loss between dense smoke and clear regions. \emph{Semantic Rectification} uses spatially aligned pixel features to rectify the tokens of a frozen foundation encoder where smoke corrupts them, before local cross-attention fuses them with pixel features. No smoke mask or transmission estimate is needed at inference, and public dehazing serves only as an auxiliary task. Our contributions are:
\begin{itemize}
\item \textbf{A mask-free framework for fire smoke removal.} FireDesmoke learns where to restore from training data and semantic priors that both follow the real smoke distribution, through Real Smoke Transfer and Semantic Rectification. Among the compared methods, FireDesmoke trained on the SmokeBench training split alone achieves the highest PSNR on that benchmark, and the full model achieves the highest zero-shot PSNR on RealX3D and the highest mean PSNR over five public dehazing test sets.
\item \textbf{Controlled ablations of each component.} With matched training budgets, the transferred pairs improve dense- and medium-smoke PSNR by 0.74 and 0.48 dB over SmokeBench pairs alone (\cref{tab:arms}), the real smoke layout outperforms a histogram-matched procedural layout and a learned haze transfer fitted to the same recordings, and Semantic Rectification preserves the clear regions: without it, they fall 1.31 dB below the input.
\item \textbf{The first paired dataset of real fire smoke for image restoration.} To the best of our knowledge, our dataset, RealFireSmoke, is the first of its kind: a fire of mixed materials burns in the scene, controlled by firefighters from ignition to extinguishment, and clear frames of the same fixed view serve as references, with evaluation masks and transmission-based regions. Its training recordings provide the smoke fields, and its held-out test recordings form a benchmark that jointly evaluates smoke removal, preservation of visible content, and content changes, revealing failures that restoration metrics alone miss. On it, FireDesmoke gains 3.73 dB in dense smoke and is the only compared method to gain at least 1 dB without lowering clear-region PSNR.
\end{itemize}

\section{Related Work}
\label{sec:related}
\noindent\textbf{Content fidelity of restoration.} With generative restoration, the main failure has shifted from missing details to hallucinated content. \emph{How far}~\cite{rw_x_howfar} and LL-Bench~\cite{rw_x_llbench} evaluate missing, replaced, or inserted objects, and LL-Bench finds that large generative models hallucinate more than conventional restoration models; VHA~\cite{rw_x_vha} and Hallucination Score~\cite{rw_x_hscore} measure altered content in super-resolution, and ProDehaze~\cite{rw_x_prodehaze} reduces hallucination in diffusion dehazing. Under real smoke, multi-view methods report color and texture changes from generative preprocessing~\cite{rw_sm44,rw_x_mllmgs}, and FireNav~\cite{rw_x_firenav} argues, in simulated fire navigation, that restoration can corrupt task-critical cues. None of these works evaluates content changes in single-image restoration of real fire smoke.

\begin{table}[t]\color{black}\centering\scriptsize\setlength{\tabcolsep}{2.2pt}\renewcommand{\arraystretch}{1.0}
\caption{\FDMcolor \textbf{Recorded smoke and haze data.} To the best of our knowledge, only ours pairs real fire smoke with recorded clear references for restoration (Table~S19). $^\dagger$FIgLib's frames before ignition are its no-smoke detection class.}\label{tab:data}
\begin{tabular}{@{}lll@{}}\toprule
Data & Smoke or haze source & Recorded reference \\\midrule
REVIDE~\cite{revide}, RealX3D~\cite{realx3d} & haze / smoke machine & yes, repeated path \\
SmokeBench~\cite{smokebench}, Desmoke6k~\cite{rw_y02} & machine (+ pellets~\cite{smokebench}) & yes, same view \\
Real test images of~\cite{desmokenet,hadn} & web / not stated & no \\
Corridor tests of~\cite{rw_sm02} & ducted heptane smoke & no \\
SmokeSeer~\cite{smokeseer} & firefighting scenes & no (rendered) \\
FLAME~\cite{rw_sm22}, FIgLib~\cite{figlib} & burns and wildfires & no$^\dagger$ \\
\textbf{RealFireSmoke} (ours) & real fire, mixed materials & yes, same view \\
\bottomrule\end{tabular}
\end{table}

\noindent\textbf{Desmoking and smoke data.} Laparoscopic desmoking~\cite{wu2024desmoke,rw_sm13,rw_sm78} has real but misaligned pairs. For fire and surveillance scenes, training smoke is mostly synthetic: DHSGAN~\cite{dhsgan} uses depth-synthesized fog/smoke pairs, DesmokeGAN~\cite{rw_sm01} rendered smoke, SDANet~\cite{rw_sm06} simulated building fires, and DesmokeNet~\cite{desmokenet} and CIANet~\cite{cianet} synthetic smoke pairs. Retinex and dark-channel pipelines~\cite{rw_sm02,rw_sm04} enhance images without training and are evaluated without a clear reference. HADN~\cite{hadn} shares the goal of removing smoke without over-enhancing clear regions; it is trained on synthetic smoke, states that real paired images are infeasible to collect, and is evaluated on 495 unpaired real images. Recorded data (\cref{tab:data}) either pair clear references with generated smoke or haze, which can be repeated, or show real smoke without a paired clear reference. The paired datasets are SmokeBench, with 9{,}975 surveillance pairs (9{,}875 for training) on which PGDR~\cite{rw_x_pgdr} and DVANet~\cite{rw_se07} report results, Desmoke6k, RealX3D, and the indoor video dataset REVIDE (see also~\cite{jin2022fog,yamaguchi2021video}). The real test images and corridor tests of earlier methods, the detection datasets FLAME~\cite{rw_sm22,rw_sm24} and FIgLib, and SmokeSeer, which renders its references from a smoke-free reconstruction and reports no quantitative results on real scenes, provide no paired clear reference for restoration. To the best of our knowledge, none pairs real fire smoke (the smoke of a fire burning in the recorded scene) with directly recorded clear frames of the same view for restoration (54 works examined: Table~S19).

\noindent\textbf{Real-world dehazing, our auxiliary task.} Dehazing models learn depth-dependent, homogeneous haze from synthetic pairs~\cite{dehazeformer,mbtaylor} or adapt to real haze~\cite{ridcp,corun,coa,bilalora,dehazesb,pglnet}.

\noindent\textbf{Degradation synthesis and haze transfer.} Depth-based~\cite{reside,narasimhan2002}, phenomenological~\cite{ridcp}, and generative~\cite{hazegen,genhaze} synthesis produce hazy training pairs from clean images. HazeFlow~\cite{hazeflow} generates non-homogeneous haze by Markov-chain Brownian motion. YOLY~\cite{yoly}, MDST-Dehaze~\cite{mdst}, and domain-exchange methods~\cite{rw_h26} transfer haze to other clean images; video matting~\cite{rw_r01}, WildSmoke~\cite{wildsmoke}, and SFMattingNet~\cite{sfmatting} composite matted or extracted real smoke; PRISM~\cite{prism} reconstructs spatially varying atmospheric light and transmission with closed-form updates and a learned transmission refinement, FujinSplat~\cite{fujinsplat} fits global RAW color corrections on real smoke pairs and applies their inverse to external clean images, and FFS-CycleGAN~\cite{ffscyclegan} synthesizes forest-fire smoke with spatially varying atmospheric light for unpaired training. The closest works transfer a learned or estimated degradation onto other clean images, for haze and weather: PHATNet~\cite{phatnet}, HTFA-Net~\cite{htfanet}, WTNet~\cite{rw_lwt01}, Di~\etal~\cite{di2026twostage}, and Jiang~\etal~\cite{rw_y01} (Sec. S8.3). Real fire smoke is more demanding: it is local and dense, and its spatial layout is essential. Real Smoke Transfer therefore calibrates smoke fields analytically on paired real-fire recordings and preserves their image-space layout when transferring them onto public clean scenes. Fitted to the same recordings, PHATNet's haze transfer places 30\% of its degradation inside the smoke region, vs.\ 68\% for our smoke fields (Sec. S8); with the network and training budget fixed, our transferred pairs outperform it and histogram-matched procedural smoke in dense smoke (\cref{tab:arms}).

\noindent\textbf{Foundation features and objectives.} Foundation features guide restoration by fusion, modulation, or routing~\cite{dinoir,peer_perceiveir,pixrestore,rw_ai31,rw_se07} or constrain the restored output~\cite{dinoir,jin2022fog}, and degraded features have been aligned with or restored toward clean ones, for restoration in DSGIR~\cite{dsgir}, robust DA-CLIP~\cite{dacliprobust}, and UniRestore~\cite{unirestore}, and for segmentation in RobustSAM~\cite{rw_se01}, which also reports dehazing and deblurring gains, and GleSAM~\cite{rw_x_glesam} (Secs. S8.1 and S8.3). Under real fire smoke, these features remain intact outside the smoke region and are corrupted inside it, so the correction must be local and smoke-aware. Semantic Rectification keeps the foundation encoder frozen and uses spatially aligned pixel features to rectify its multi-layer patch tokens before local fusion and RGB reconstruction; without rectification, the tokens yield a similar dense-smoke PSNR but degrade clear content, even far from the smoke (\cref{sec:mech}). Region-selective restoration usually relies on a predicted or specified region~\cite{rw_se03,rw_sm51,rw_sm38,rw_h09,rw_se18,desmokenet,hadn}; FireDesmoke has no mask input or mask branch (more related work: Sec. S8.3).

\section{Methodology}
\label{sec:method}
\begin{figure*}[t]\FDMcolor 
\centering
\includegraphics[width=\textwidth]{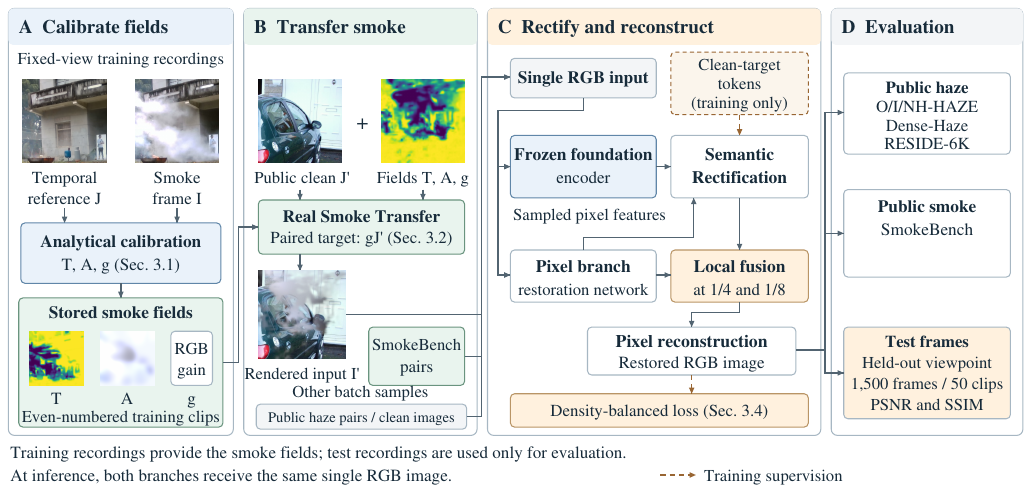}
\caption{\FDMcolor Overview of FireDesmoke. (A) Training recordings and their same-view clear references provide only the smoke fields $T$ (effective transmission), $A$ (atmospheric light), and $g$ (RGB exposure gain) (\cref{sec:fields}). (B) Real Smoke Transfer renders the fields onto Open Images~\cite{openimages} scenes, giving exposure-matched training pairs used with SmokeBench pairs, public haze pairs, and clean images (\cref{sec:transfer}). (C) Semantic Rectification: sampled pixel features guide the rectification of the foundation encoder's input tokens, which are fused with pixel features at $1/4$ and $1/8$ resolution; clean-target tokens supervise it (\cref{sec:carrier,sec:loss}); no mask or transmission is used at inference. (D) Held-out test recordings and public test splits are used only for evaluation (\cref{sec:exp}).}
\label{fig:pipeline}
\end{figure*}
\noindent\textbf{Overview.} FireDesmoke enforces one requirement, that both the training data and the semantic priors follow the real smoke distribution, in one pipeline of three stages that decouples the acquisition of smoke degradation from that of scene content (\cref{fig:pipeline}). \emph{Calibration}: fixed-view recordings of real fire are converted into smoke fields, the only information taken from the fire scene (\cref{sec:fields}). \emph{Transfer}: the fields are rendered onto public clean images, which provide the scene content of the training pairs, and combined with SmokeBench pairs (\cref{sec:transfer}). \emph{Rectification and reconstruction}: a single RGB image is fed to the pixel branch and to a frozen foundation encoder, whose tokens are rectified with local pixel features where smoke corrupts them and fused with the reconstruction features at two resolutions (\cref{sec:carrier}). The calibrated transmission is used throughout: it renders the pairs, guides crop sampling, balances the loss, masks the token supervision (Sec. S1.2), and defines the evaluation regions; it is never used at inference.

\subsection{Smoke Field Calibration}
\label{sec:fields}
\begin{figure}[t]
\centering
\includegraphics[width=\columnwidth]{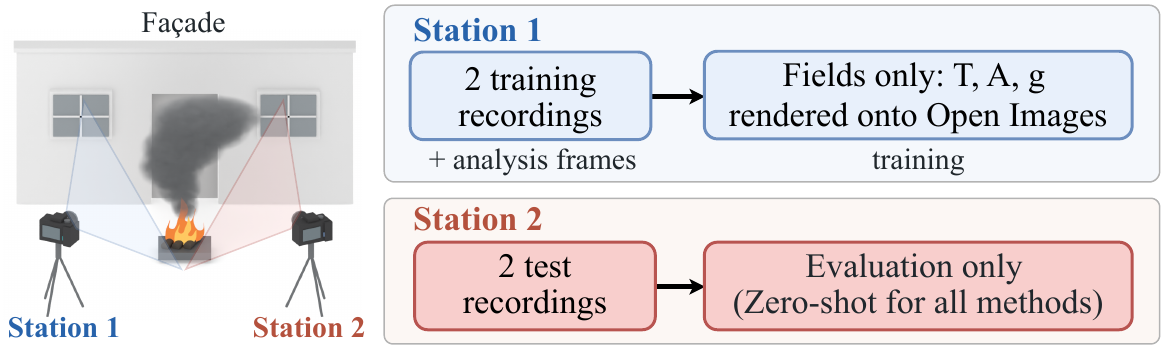}
\caption{\textbf{Capture and data roles.} Real fires of common materials, operated by the fire department, are recorded from two fixed viewpoints of one fa\c{c}ade in turn, so clear frames of the same view serve as references. Station-1 training recordings provide only the smoke fields $T$, $A$, and $g$, rendered onto Open Images scenes for training, plus held-out analysis frames; Station-2 test recordings are used only for evaluation, zero-shot for all methods.}
\label{fig:capture}
\end{figure}

\noindent\textbf{Acquisition.} Our recordings combine real fires of common fire-scene materials, controlled operation, and same-view clear references (\cref{fig:capture}). They were captured at the fire department's search-and-rescue training ground on a dry day. The fire department burned wood, wood chips, and paper, then built a larger fire by adding wood, dry and wet grass, discarded clothing, and gasoline once the fire had died down or was judged safe. Four firefighters with a fire engine and a water-spraying quadruped robot lit, tended, and extinguished the fire under a supervising officer; the fire department decided the fuel, its placement, when more could be added, and our camera positions and distance. Fixed cameras recorded the outdoor fire from two viewpoints (Stations~1 and~2) in turn, so clear frames of the same view serve as references. The first 20 minutes, two recordings from Station~1, provide the smoke fields (even-numbered \mbox{20-s} clips) and held-out analysis frames (odd-numbered clips; Sec. S6.1); the next 20 minutes, two recordings from Station~2, are used only for testing (\cref{sec:exp}). Capture took 8--9 hours on site. The fire department authorized recording, staff (the firefighters and others on site) gave informed consent beforehand, and the institutional research ethics center determined that, with people removed, the study did not meet its criteria for review, so no review was required; no person was staged inside smoke, and human-shaped targets are mannequins and cut-outs.

\noindent\textbf{Calibration.} For each training recording, smoke frames $I$ are paired with a fixed-view temporal reference $J$ obtained by median aggregation in linear RGB (Sec. S1.1). We model each smoke pixel as
\begin{equation}
I = g\odot\big(T\odot J + (1-T)\odot A\big),
\label{eq:model}
\end{equation}
where $T$ is a single-channel effective transmission, $A$ a per-pixel RGB atmospheric light (the color of fully opaque smoke), and $g$ a per-frame RGB exposure gain. The gain $g$ is the ratio of the per-channel means of $I$ and $J$ over a fixed ground strip. $T$ and $A$ are refined in four rounds of four closed-form steps: a global least-squares $A$, a spatial $A$ smoothed by normalized convolution~\cite{rw_r19}, a color-based $T$, and its confidence-weighted fusion with a texture-based one (Sec. S1.1). Spatial smoothness of $A$ regularizes the decomposition into the effective quantities $T$ and $A$. We refer to the calibrated $T$, $A$, and $g$ as smoke fields, and to the training pairs rendered from them as transferred pairs.

Firelight changes the exposure from frame to frame, so $g$ is estimated per frame rather than absorbed into the smoke layer. Flames, water spray, and moving staff are masked: a pixel is marked invalid if it lies in such a dynamic mask or its sRGB reconstruction error exceeds 0.15. The smoke moves continuously, so the fields are stored per frame, and $280{\times}280$ windows are sampled from every calibrated frame of the even-numbered training clips (selection: Sec. S1.1). The stored fields are only $T$ at half resolution, $A$ at one-eighth resolution, the three components of $g$, and the invalid-pixel mask; the recorded RGB frames and their references are never used as inputs or targets in restoration training.

\subsection{Real Smoke Transfer}
\label{sec:transfer}
Real Smoke Transfer renders a window of the smoke fields onto a public clean image $J'$ (Open Images~\cite{openimages} validation and test images; crops around an annotated object, random flipping) as \begin{equation} I' = g\odot\big(T\odot J' + (1-T)\odot A\big),\qquad \text{target } g\odot J', \label{eq:render} \end{equation} so that the scene content of every transferred pair comes from Open Images (encoding: Sec. S1.1). FireDesmoke is trained with five transferred pairs and five SmokeBench pairs per batch (ten smoke pairs), together with four public haze pairs and two clean images. The data ablations in \cref{tab:arms} change only the smoke pairs. SmokeBench crops use the transmission estimated from each pair (Sec. S1.1) for the same crop sampling and density-balanced loss (\cref{sec:loss}).

For comparison, procedural pairs (\cref{tab:arms}) keep each sampled window's exact $T$ histogram, mean $A$ over smoky pixels ($T{<}0.4$), and $g$, but replace the transmission layout with rank-mapped four-octave noise and the spatial $A$ with its smoky-pixel mean. PHATNet pairs are obtained by fitting PHATNet's transfer network to the same training recordings and transferring the same windows onto the same public clean images (Sec. S8). Identity pairs replace a smoke pair with a clean pair whose input equals its target. Table S1 lists the smoke pairs of each setting.

\noindent\textbf{Content leakage check.} On 400 training windows of transferred pairs, fire-scene fine structure explains $R^2{=}0.004$ of the rendered input's fine structure in dense smoke (unrelated scenes: 0.002; 10\% injected fire-scene content: 0.375; Sec. S3): the transfer carries no measurable fire-scene structure.

\subsection{Semantic Rectification}
\label{sec:carrier}
Fire smoke makes the reliability of foundation features spatially non-uniform: clear-region tokens remain close to their clean counterparts, whereas dense-smoke tokens deviate strongly. Fused without correction, they make the model alter clear content far from the smoke (\cref{sec:mech}). We therefore rectify the tokens of the degraded input, conditioned on pixel features, before they guide RGB reconstruction. The pixel branch restores the image; in parallel, the foundation branch, a frozen foundation encoder, extracts patch tokens from four of its blocks. The rectification module, two 128-dimensional local transformer blocks, uses sampled, spatially aligned pixel features to jointly predict residual updates for the four token layers. The rectified tokens are concatenated and projected to a shared 128-dimensional map, which is resampled to the $1/4$ and $1/8$ grids and provides keys and values for local cross-attention, with pixel features as queries, so each pixel attends only to nearby tokens. The residual updates and the fusion projections are zero-initialized, so training starts from the output of the pixel branch. Cosine and smooth-L1 losses supervise the rectified tokens with the clean target's tokens (Sec. S1.2).

\noindent\textbf{Implementation.} The pixel branch is AdaIR~\cite{peer_adair}, initialized from the authors' dehazing weights, and the foundation encoder is DINOv3 ViT-L/16~\cite{dinov3}, read at blocks 6, 12, 18, and 24. The foundation encoder and the AdaIR encoder are frozen, and the 12.7M decoder, rectification, and fusion parameters are updated for 10k steps with a batch size of 16 using AdamW; training takes 3.6 hours on one NVIDIA H200 GPU. The network has 334.0M parameters, 303.2M of them in the frozen foundation encoder. The methods compared on SmokeBench are trained on its training split, so for that benchmark FireDesmoke is trained on that split alone (FireDesmoke$^\ast$; Sec. S1.3).

\subsection{Training Objective}
\label{sec:loss}
Smoke pairs use L1 plus 0.05 Sobel-L1 losses. Their pixels are grouped into four effective-transmission ranges, $[0,.2)$, $[.2,.5)$, $[.5,.95)$, and $[.95,1]$, using the rendered or, for SmokeBench, the estimated transmission; ranges with at least 256 pixels are averaged equally. This density-balanced loss, $L_{\mathrm{smoke}}$, prevents the abundant clear pixels from dominating the supervision of dense smoke ($+$0.28 dB in dense smoke over a pixel-averaged loss, Table S6; cf.~\cite{rw_h32}).
The total loss is
\begin{equation}
L = L_{\mathrm{smoke}} + L_{\mathrm{pub}} + 0.1L_{\mathrm{clean}} + 0.1(Z_s+Z_p),
\label{eq:loss}
\end{equation}
where $L_{\mathrm{pub}}$ is the pixel average of the same L1 and Sobel-L1 terms over the public haze pairs, $L_{\mathrm{clean}}$ the L1 loss on the clean images, and $Z_s$ and $Z_p$ the rectification losses (cosine distance plus 0.1 smooth-L1) on smoke pairs and on haze pairs and clean images.

\section{Experiments}
\label{sec:exp}
\begin{table}[t]\color{black}\centering\footnotesize\setlength{\tabcolsep}{2.4pt}\renewcommand{\arraystretch}{1.0}
\caption{\FDMcolor \textbf{Quantitative comparison on our test set} (1,500 frames / 50 clips, Station~2, zero-shot for all methods): masked RGB PSNR (dB) in dense ($T{<}0.2$), medium ($0.2{\le}T{<}0.5$), and clear ($T{\ge}0.9$) regions of the estimated transmission, on annotated static objects, and overall (all valid pixels of the evaluation bands), with overall SSIM. Released weights; $^\ast$: trained on the SmokeBench training split alone, the baselines by us with their published settings. Listed: the five of the 116 baseline models with the highest dense-smoke PSNR (DehazeSNN-L: RS-Haze weights; HistoFusionNet: NH-HAZE weights); two of them, DiffDehaze and RealRestorer, generative models that add content absent from the input (\cref{fig:teaser}), have their own block. FireDesmoke$^\ast$: \cref{sec:carrier}; the last row is the full model. All 116 models: Table S2; sources: Table \SRCTAB. \rankfirst{Best}/\ranksecond{second} among the rows shown; input and that block excluded.}\label{tab:main}
\begin{tabular}{@{}l c c c c cc@{}}\toprule
\multirow{2}{*}{Method} & Dense & Medium & Clear & Object & \multicolumn{2}{c}{Overall} \\
\cmidrule(lr){2-2}\cmidrule(lr){3-3}\cmidrule(lr){4-4}\cmidrule(lr){5-5}\cmidrule(lr){6-7}
 & PSNR & PSNR & PSNR & PSNR & PSNR & SSIM \\ \midrule
Input & 15.16 & 18.07 & 30.52 & 26.61 & 25.03 & 0.881 \\
\midrule
DehazeSNN-L~\cite{rw_x_dehazesnn} & 17.44 & 18.95 & 20.10 & 20.61 & 19.48 & 0.786 \\
HistoFusionNet~\cite{rw_x_histofusion} & 17.36 & 19.81 & 24.30 & 22.39 & 22.11 & 0.806 \\
DehazeDCT~\cite{rw_h07} & 17.09 & 19.87 & \ranksecond{26.57} & 22.73 & 23.37 & 0.832 \\
\midrule\multicolumn{7}{@{}l}{\textit{Trained on the SmokeBench training split}} \\
Uformer-B$^\ast$~\cite{uformer} & 16.78 & 18.50 & 21.79 & \ranksecond{24.19} & 20.34 & 0.856 \\
PGDR$^\ast$~\cite{rw_x_pgdr} & 16.66 & 17.39 & 20.43 & 17.88 & 19.34 & 0.810 \\
DVANet$^\ast$~\cite{rw_se07} & 18.36 & \ranksecond{19.95} & 25.86 & 23.19 & \ranksecond{23.47} & \ranksecond{0.866} \\
FireDesmoke$^\ast$ (ours) & \ranksecond{18.37} & 19.91 & 24.88 & 23.25 & 22.86 & 0.863 \\
\midrule\multicolumn{7}{@{}l}{\textit{Generative models that add content (not ranked)}} \\
DiffDehaze~\cite{hazegen} & 16.97 & 19.23 & 19.59 & 18.53 & 18.60 & 0.513 \\
RealRestorer~\cite{rw_l202} & 20.86 & 22.15 & 25.21 & 24.53 & 23.41 & 0.814 \\
\midrule
\textbf{FireDesmoke} (ours) & \rankfirst{18.89} & \rankfirst{20.70} & \rankfirst{30.93} & \rankfirst{28.23} & \rankfirst{26.53} & \rankfirst{0.887} \\
\bottomrule\end{tabular}\end{table}
\begin{figure*}[t]\FDMcolor 
\centering
\includegraphics[width=\textwidth]{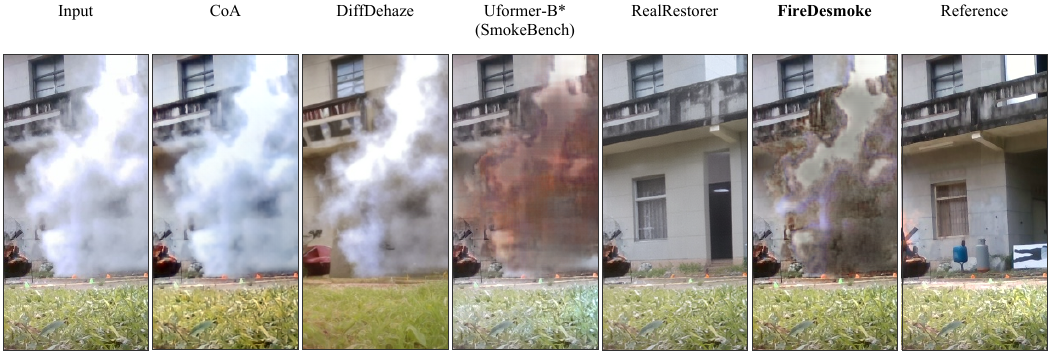}
\caption{\FDMcolor \textbf{Qualitative comparison on a test frame selected by dense-smoke coverage.} CoA~\cite{coa} retains the smoke and shifts its color; DiffDehaze reduces it but changes the colors of the scene; Uformer-B$^\ast$ (trained on SmokeBench) leaves dark residue and darkens the building. RealRestorer produces a clean fa\c{c}ade, but the blue drum, gas cylinder, and cut-out board behind the smoke in the reference are missing or altered; FireDesmoke reduces the bright smoke while preserving details in the clear foreground. The Reference panel combines the references (exposure-matched clear views) of the evaluation bands. Figure~S3 shows the evaluation masks and dense-smoke samples.}
\label{fig:qual}
\end{figure*}
\noindent\textbf{Evaluation protocol.} Station~2 provides 1,500 test frames in 50 clips of 20 s (30 fps, $640\times480$, 30 evaluated frames per clip). Staff in these frames are replaced by inpainted background before evaluation, and the replaced pixels are excluded (Sec.~S6.1). The test recordings are zero-shot for every method, including FireDesmoke: none of their images is used for training or adaptation, and they contribute no smoke fields to training. Public dehazing test images are not used to train or validate FireDesmoke. Every setting of FireDesmoke is trained for a fixed number of steps.

\noindent\textbf{Metrics.} Metrics are computed in RGB unless a dataset has its own evaluation protocol (SmokeBench, \cref{sec:ext}). We evaluate on three partially overlapping vertical bands, one per group of static objects (the evaluation bands), whose union covers 37\% of the image width; ``overall'' denotes their valid pixels. The reference is a clear frame of the same view from another clip of the same recording, exposure-matched to the input and shared across methods (selection and evaluation masks: Sec. S6.2). We report masked PSNR in regions of estimated effective transmission (dense $T{<}0.2$, medium $[.2,.5)$, clear $[.9,1]$) and on annotated static objects, and PSNR and SSIM ($7{\times}7$ windows inside the evaluation mask) over all valid pixels. Results are averaged over frames per clip, clips per recording, then equally over recordings and bands; per-recording differences are in Secs. S2.2--S2.3. Differences and dataset means are computed from unrounded values.

\noindent\textbf{Baselines.} We compare with \NEXT{} baseline models for dehazing, desmoking, and all-in-one restoration, using the authors' released weights and default settings (sources: Table \SRCTAB). Models marked with $^\ast$ are trained on the 9,875 SmokeBench training pairs alone, the baselines by us: Uformer-B, MB-TaylorFormer-B~\cite{mbtaylor}, PGDR, and DVANet with their published settings and, for the last two, the authors' code and default settings. We also train PGL-Net~\cite{pglnet} after removing the public haze test images from its training set and adapt PHATNet to our training recordings ($^\diamond$ in Table S2). All methods use the same test inputs and evaluation code.

\begin{table*}[t]\color{black}\centering\footnotesize\setlength{\tabcolsep}{5pt}\renewcommand{\arraystretch}{1.0}
\caption{\FDMcolor \textbf{Ablation study with matched training budgets} on our test set (masked PSNR, dB; regions as in \cref{tab:main}). Every row is FireDesmoke; the first column states what differs from the full model. All settings share the network (unless stated), initialization, batch composition (ten smoke pairs, four public haze pairs, two clean images), density-balanced loss (pixel average when all ten are identity pairs), and training budget; the SmokeBench-only setting thus keeps the haze pairs and clean images and is not FireDesmoke$^\ast$ (\cref{tab:main}). Smoke pairs per batch (\cref{sec:transfer}): R transferred (Real Smoke Transfer), G procedural, P PHATNet, N identity, SB SmokeBench. Last column: first row of the block minus row, dense-smoke PSNR. \rankfirst{Best}/\ranksecond{second} per column within each block; ties share the rank.}\label{tab:arms}
\begin{tabular}{l l c c c c c r}\toprule
Setting & Smoke pairs & Dense & Medium & Clear & Object & Overall & $\Delta$ dense PSNR \\\midrule
\multicolumn{8}{l}{\textit{Smoke data, Semantic Rectification, and the foundation branch ($\Delta$: full model $-$ row)}} \\
\textbf{Full model} & 5\,R + 5\,SB & \rankfirst{18.89} & \rankfirst{20.70} & 30.93 & \ranksecond{28.23} & \rankfirst{26.53} & -- \\
Procedural smoke + SmokeBench & 5\,G + 5\,SB & 18.44 & 20.38 & \rankfirst{30.96} & 28.10 & 26.36 & +0.45 \\
PHATNet transfer + SmokeBench & 5\,P + 5\,SB & 18.16 & 20.20 & 30.16 & 28.06 & 26.11 & +0.74 \\
Identity pairs + SmokeBench & 5\,N + 5\,SB & 17.26 & 19.64 & 30.69 & 27.88 & 25.99 & +1.64 \\
SmokeBench only & 10\,SB & 18.15 & 20.21 & \ranksecond{30.94} & \rankfirst{28.50} & \ranksecond{26.40} & +0.74 \\
w/o Semantic Rectification & 5\,R + 5\,SB & \ranksecond{18.74} & \ranksecond{20.51} & 29.21 & 26.86 & 25.66 & +0.15 \\
Pixel branch only & 5\,R + 5\,SB & 17.32 & 19.59 & 26.25 & 25.10 & 23.20 & +1.58 \\
\midrule
\multicolumn{8}{l}{\textit{A single type of smoke pairs ($\Delta$: Real Smoke Transfer only $-$ row)}} \\
\textbf{Real Smoke Transfer only} & 10\,R & \rankfirst{17.14} & \rankfirst{19.91} & \ranksecond{30.49} & \rankfirst{27.48} & \rankfirst{25.87} & -- \\
Procedural smoke only & 10\,G & \ranksecond{16.78} & \ranksecond{19.50} & 30.29 & \ranksecond{27.15} & 25.59 & +0.36 \\
PHATNet transfer only & 10\,P & 16.75 & 19.36 & 30.39 & 27.14 & \ranksecond{25.75} & +0.39 \\
Identity pairs only & 10\,N & 15.16 & 18.07 & \rankfirst{30.52} & 26.61 & 25.03 & +1.98 \\
\bottomrule\end{tabular}\end{table*}

\subsection{Comparison with State-of-the-Art Methods}
\label{sec:external}
\Cref{tab:main} and \cref{fig:teaser}d evaluate two requirements on the same 1,500 test frames: restoration in dense and medium smoke, and preservation of the already visible scene. FireDesmoke improves dense- and medium-smoke PSNR by 3.73 and 2.63 dB over the input. In clear regions the requirement is to remain at the level of the input: FireDesmoke changes them by $+$0.41 dB over both recordings, $+$0.85 dB in one and $-$0.03 dB in the other. No other compared method satisfies both requirements: each of the nine of the 116 baseline models that gain at least 1 dB in dense smoke falls at least 3.9 dB below the input in clear regions, generative or not (Sec. S9; \cref{fig:teaser}d). The same holds at a finer scale: taking one clip in one evaluation band as a unit, 95 units contain both dense-smoke and clear pixels, and FireDesmoke increases dense-smoke PSNR without lowering clear-region PSNR in 56 of them, vs.\ at most 11 for any method of Table S15 with a dense-smoke gain of at least 1 dB. FireDesmoke has the highest dense- and medium-smoke PSNR of all compared methods except RealRestorer, a general-purpose generative restoration model (20.86 and 22.15 dB), which hallucinates people and removes a visible flame (\cref{fig:teaser}; Sec. S9) and loses 5.30 dB in clear regions. The four methods of the SmokeBench block show the same pattern: all, including FireDesmoke$^\ast$, gain 1.5--3.2 dB in dense smoke but fall 4.7--10.1 dB below the input in clear regions. On annotated static objects, including mannequins and cut-outs, FireDesmoke reaches 28.23 dB, vs.\ 26.61 dB for the input, 26.65 dB for the best baseline model (Table S2), and 24.53 dB for RealRestorer. Its overall PSNR and SSIM, 26.53 dB and 0.887, are also the highest.

\noindent\textbf{Hallucinated content.} The clear-region losses above are changes to content the input shows; in the worst case, a model generates content the input does not show, and a person detector applied to these outputs reports people where the input has none. We select output person detections that match no input detection and lie in dense or medium smoke, then inspect every candidate (Sec. S9.4). Over all 1,500 test frames, this analysis finds no person hallucination for FireDesmoke (no candidate at all), 16 confirmed for DiffDehaze, 7 for CoDehaze~\cite{opp5_codehaze}, and 2 for RealRestorer (Table \PERSONTAB): RealRestorer turns a gas cylinder hidden in smoke into a standing person and adds another at a window, both in one frame (\cref{fig:teaser}a), and DiffDehaze generates a person in smoke (\cref{fig:teaser}b). These methods replace objects (\cref{fig:qual}, Fig. \HALLFIG{}), and RealRestorer removes a flame (\cref{fig:teaser}c).

\subsection{Effect of Real Smoke Transfer}
\label{sec:fields_exp}
\noindent\textbf{Real smoke in the full model.} Baselines trained on public haze or smoke data fail on real fire smoke (\cref{tab:main}). With the network, the remaining public data, and the budget fixed, using transferred pairs in place of half of the SmokeBench pairs adds 0.74 dB in dense smoke and 0.48 dB in medium smoke over using SmokeBench pairs for all ten smoke pairs, with clear regions unchanged (\cref{tab:arms}); replacing the transferred pairs with identity pairs costs 1.64 dB in dense smoke, in both recordings alike. \textbf{Real smoke layout.} Procedural smoke with the same transmission histogram fragments the smoke region (median 25 connected components of $T{<}0.5$ per window, vs.\ 6; Table S4) and costs 0.45 dB in dense smoke and 0.17 dB overall, with clear regions unchanged; both margins remain positive in each recording and under every alternative reference and band definition (Sec. S2.3). With a single type of smoke pairs and $A$ set to its smoky-pixel mean in both, the real layout reaches 17.61 dB in dense smoke vs.\ 16.78 dB for the procedural layout; with the calibrated spatial $A$ in both, 17.14 vs.\ 17.04 dB (Table S5). PHATNet transfer places only 30\% of its degradation inside the smoke region and degrades the clear part of training windows 16 times as much as Real Smoke Transfer (Sec. S8); the model trained on it is lower than FireDesmoke by 0.74 dB in dense smoke, 0.77 dB in clear regions, and 0.42 dB overall, again in each recording, and is 0.36 dB below the input in clear regions.

\noindent\textbf{Real Smoke Transfer as the only smoke data.} With no SmokeBench pair in training, transferred pairs alone still raise dense-smoke PSNR by 1.98 dB over the input with clear regions within 0.04 dB of it, outperforming procedural and PHATNet pairs (\cref{tab:arms}).

\subsection{Effect of Semantic Rectification}
\label{sec:mech}
\noindent\textbf{Semantic Rectification preserves clear content when training on real smoke.} Without it, i.e., with the frozen input tokens fused directly, the model is only 0.15 dB lower in dense smoke but loses 1.72 dB in clear regions (in all 50 clips), 1.37 dB on static objects, and 0.87 dB overall; in clear regions it is 1.31 dB below the input. Setting only the weight of the rectification loss to zero keeps the clear-region PSNR and instead lowers dense-smoke SSIM from 0.426 to 0.413: the rectification module preserves the clear scene, and the rectification loss improves structure where smoke is dense. Real Smoke Transfer and Semantic Rectification are complementary: without rectification, replacing half of the SmokeBench pairs with transferred pairs raises dense-smoke PSNR by 1.13 dB but lowers clear-region PSNR by 1.62 dB; with it, the 0.74 dB dense-smoke gain leaves clear regions unchanged (Table S6). Rectification raises the dense-smoke cosine similarity of the input tokens to those of the clear reference by 0.09--0.14 across the four blocks, while clear-region similarity changes by at most 0.005 (Sec. S8). Without the foundation branch, the pixel branch trained on the same batches is 4.27 dB below the input in clear regions (\cref{tab:arms}).

\noindent\textbf{Changes are localized to the smoke.} Fig. \LOCFIG{} groups clear pixels by their distance to the nearest dense-smoke pixel. Pooled over the recordings, FireDesmoke remains above the input at every distance, by 0.15--1.23 dB within 128\,px of the smoke, and its changes decrease with distance: the mean absolute change falls from 0.037 next to the smoke to 0.007 beyond 128\,px, where PSNR remains 0.08 dB above the input. Without rectification, the model is below the input at every distance, by 2.09 dB beyond 128\,px, with 1.5--2.6 times the change magnitude of FireDesmoke.

\subsection{Results on Public Benchmarks}
\label{sec:ext}
\noindent\textbf{Public smoke datasets.} PGDR and DVANet, recent methods that report SmokeBench results, release no SmokeBench weights, so we retrain both on its training split with the authors' code, train FireDesmoke on the same split (FireDesmoke$^\ast$), and evaluate all with the official SmokeBench evaluation code. FireDesmoke$^\ast$ achieves the highest Y-channel PSNR, 30.47 dB, above PGDR$^\ast$ (29.97 dB), DVANet$^\ast$ (29.61 dB), and the benchmark's released outputs, and the second-best SSIM and LPIPS, above DVANet$^\ast$ on all three (Table \SMOKETAB; RGB PSNR: Sec. S4). On the indoor machine smoke of RealX3D, zero-shot for all methods, FireDesmoke obtains 18.19 dB RGB PSNR, above every baseline model (next: GenDeg~\cite{peer_gendeg}, 17.94 dB; Table S8). Machine smoke, however, differs from fire smoke: in both datasets it is a nearly full-frame layer ($T{<}0.9$ coverage 0.96 and 1.00, vs.\ 0.51 in smoky crops of our test recordings; Table S4). The three methods with the highest SmokeBench PSNR, each trained on that split alone, lose 4.7--10.1 dB in clear regions of our test recordings (\cref{tab:main}).

\noindent\textbf{Public dehazing (auxiliary).} Under a common resized RGB protocol, FireDesmoke, whose haze pairs come from the training splits of O-HAZE, I-HAZE, NH-HAZE, NH-HAZE21, Dense-Haze, and RESIDE-6K, achieves an unweighted mean PSNR of 22.87 dB, higher than all baseline models (next: GenDeg, 22.43 dB; Table \HAZETAB{}).

\section{Conclusion}
\label{sec:concl}
FireDesmoke restores only where the smoke is: Real Smoke Transfer provides training pairs with the spatial layout of real fire smoke, and Semantic Rectification rectifies the foundation features that smoke corrupts. Controlled ablations show that the two are complementary: the real smoke layout improves dense-smoke restoration, and Semantic Rectification lets this gain come without lowering the clear regions. On the 1,500 test frames, FireDesmoke gains 3.73 dB in dense smoke with clear regions kept at the input level on average, the only compared method to gain at least 1 dB there without lowering clear-region PSNR. A higher PSNR cannot compensate for hallucinated people or a removed flame: useful smoke removal must preserve the evidence in the scene. RealFireSmoke (test frames with staff replaced, references, masks, evaluation results, and smoke fields) will be available by camera-ready through a data-use application and public after data cleaning; the code will be public.

\section*{LLM Usage Statement}
We used Anthropic Claude (Claude Code), OpenAI Codex and OpenAI ChatGPT as assistants throughout this work: to search for and check references, to refine the writing, to help write code, to set up experiment configurations, and to check experimental results. The research ideas and the method are the authors' own. The authors made all research decisions, verified the references against the original publications and the reported numbers against the experiment outputs, and take full responsibility for the paper.

{
\small
\bibliographystyle{ieeenat_fullname}
\bibliography{main}
}
\end{document}